\documentclass{article}

\PassOptionsToPackage{numbers, compress}{natbib}

 \usepackage[dblblindworkshop, final]{neurips_2025}

\workshoptitle{AI4Mat: AI for Accelerated Materials Design}

\usepackage[utf8]{inputenc} 
\usepackage[T1]{fontenc}    
\usepackage{hyperref}       
\usepackage{url}            
\usepackage{booktabs}       
\usepackage{amsfonts}       
\usepackage{nicefrac}       
\usepackage{microtype}      
\usepackage{xcolor}         
\usepackage{amsmath}
\usepackage{listings}
\usepackage{graphicx}
\usepackage{pgfplots}
\pgfplotsset{compat=1.18}
\usepackage{siunitx}

\title{SAM-EM: Real-Time Segmentation for Automated Liquid Phase Transmission Electron Microscopy}

\author{%
  Alexander Wang \thanks{Equal contribution.} \\
  Georgia Institute of Technology\\
  \texttt{awang652@gatech.edu} \\
  \And
  Max Xu  \footnotemark[1]\\
  Northwestern University \\
  \texttt{maxxu@u.northwestern.edu} \\
  \And
  Risha Goel \\
  Georgia Institute of Technology\\
  \texttt{rgoel63@gatech.edu} \\
  \And
  Zain Shabeeb \\
  Georgia Institute of Technology\\
  \texttt{zshabeeb3@gatech.edu} \\
  \And
  Isabel Panicker \\
  Georgia Institute of Technology\\
  \texttt{ipanicker@gatech.edu} \\
  \And
  Vida Jamali \thanks{Corresponding author. \texttt{vida@gatech.edu}} \\
  Georgia Institute of Technology\\
  \texttt{vida@gatech.edu} \\
}

\begin{document}

\maketitle

\begin{abstract}
 The absence of robust segmentation frameworks for noisy liquid phase transmission electron microscopy (LPTEM) videos prevents reliable extraction of particle trajectories, creating a major barrier to quantitative analysis and to connecting observed dynamics with materials characterization and design. To address this challenge, we present Segment Anything Model for Electron Microscopy (SAM-EM), a domain-adapted foundation model that unifies segmentation, tracking, and statistical analysis for LPTEM data. Built on Segment Anything Model 2 (SAM~2), SAM-EM is derived through full-model fine-tuning on 46,600 curated LPTEM synthetic video frames, substantially improving mask quality and temporal identity stability compared to zero-shot SAM~2 and existing baselines. Beyond segmentation, SAM-EM integrates particle tracking with statistical tools, including mean-squared displacement and particle displacement distribution analysis, providing an end-to-end framework for extracting and interpreting nanoscale dynamics. Crucially, full fine-tuning allows SAM-EM to remain robust under low signal-to-noise conditions, such as those caused by increased liquid sample thickness in LPTEM experiments. By establishing a reliable analysis pipeline, SAM-EM transforms LPTEM into a quantitative single-particle tracking platform and accelerates its integration into data-driven materials discovery and design. Project page: \href{https://github.com/JamaliLab/SAM-EM}{github.com/JamaliLab/SAM-EM}.
\end{abstract}

\section{Introduction}

In situ liquid phase transmission electron microscopy (LPTEM) is an emerging microscopy technique with an unparalleled combination of spatial and temporal resolution, enabling direct visualization of the nanoscale dynamics of materials in liquid solutions using an electron microscope~\cite{ross2015opportunities,mirsaidov2020liquid,woehl2020electron,moreno2022recent}. These features make LPTEM a powerful tool for single-particle tracking and for probing how nanoparticles nucleate, self-assemble, and transform under native conditions~\cite{jamali2021anomalous, Panicker_Shabeeb_Hargus_Jamali_2025,chen2020nucleation, luo2017quantifying, ou2020kinetic}. Looking ahead, the role of in situ electron microscopy is poised to evolve from a passive characterization technique into an active feedback-driven experimental platform~\cite{schorb2019software, kalinin2023machine, roccapriore2022automated, botifoll2022machine}. In such a paradigm, the microscope becomes not only an imaging instrument but also a controllable environment, where experimental knobs such as electron beam dose rate, ionic concentration, or fluid mixing can be modulated in real time. Coupled with on-the-fly analysis, this feedback allows researchers to actively steer nanoscale assembly and transformation, turning LPTEM into a platform for rapid hypothesis testing and iterative exploration, therefore accelerating materials discovery. Realizing this vision requires analysis pipelines that operate during acquisition, extracting particle trajectories in real time, and feeding them back to the microscope to guide automated control of experimental parameters.

\setlength{\textfloatsep}{3em}
\begin{figure*}[t]
\centering
\includegraphics[clip=true,width=\textwidth]{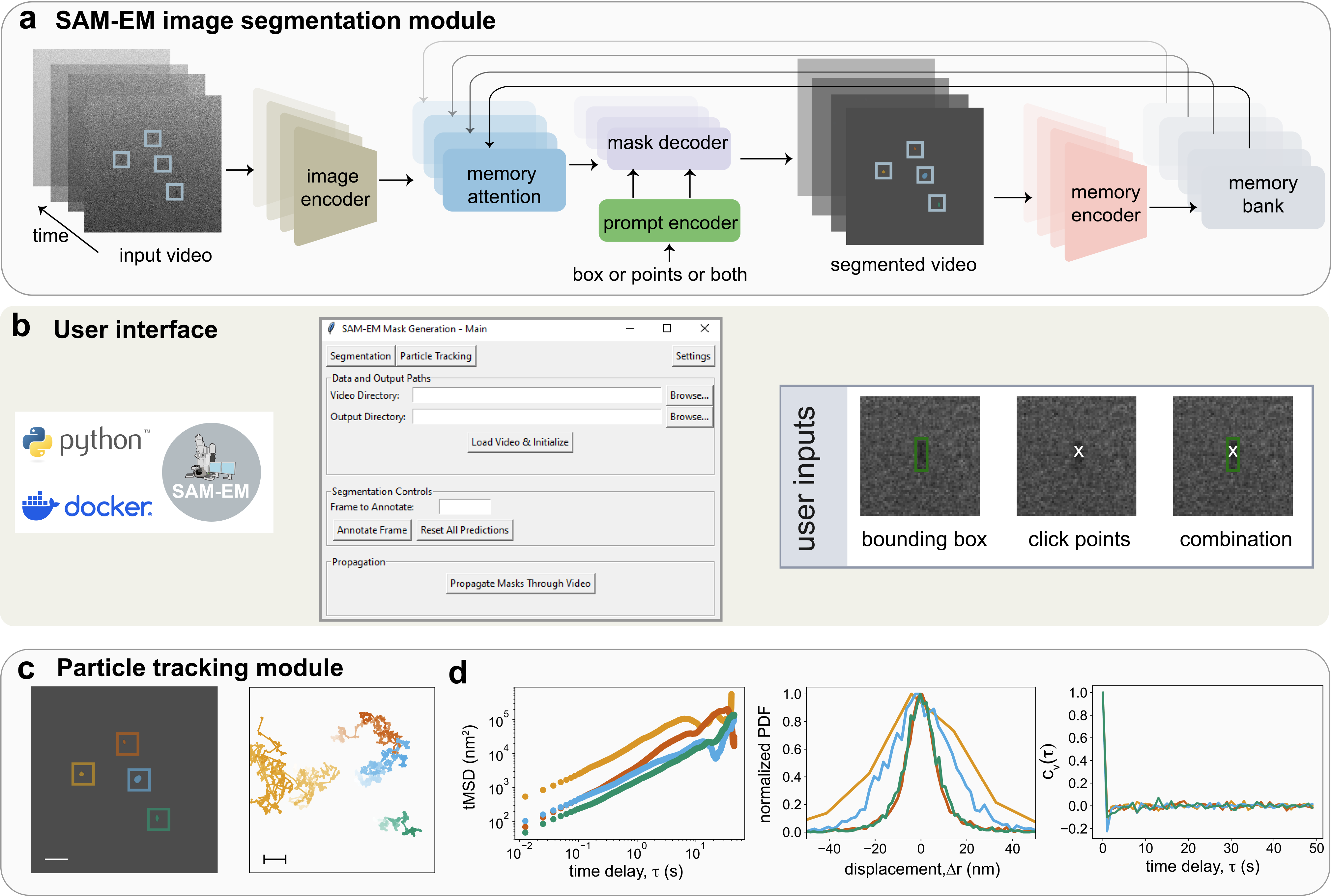}
\caption{\textbf{SAM-EM framework for particle segmentation and tracking in LPTEM.} \textbf{a}, The segmentation module segments an LPTEM video with multiple nanoparticles, each prompted once on the first frame. \textbf{b}, Deployment on any Python or Docker system with a graphical interface that enables particle selection via bounding-box prompts and statistical test selection. \textbf{c}, Extracted trajectories undergo quantitative analyses, including spatiotemporal paths (color-coded over 0-45 seconds), mean squared displacements, displacement distributions, and velocity autocorrelations. Scale bar, 250 nm.}
\label{fig:fig1}
\vspace{-1em}
\end{figure*}

Recent advances in vision foundation models suggest a path toward this goal. The Segment Anything Model 2 (SAM~2), a vision transformer trained on 50.9K videos with 93.7 million parameters, was released by Meta in 2024 as a generalization of its predecessor SAM. SAM~2 supports promptable video segmentation at real-time speeds of nearly 50 frames per second (fps), conditioning its predictions both on user prompts and on a memory bank of previously segmented frames. This design allows SAM~2 to identify and track objects that move in and out of frame or cross paths with one another, features that are critical for particle tracking in LPTEM.

However, when applied directly (zero-shot) to noisy LPTEM videos, SAM~2 fails to achieve robust performance, particularly under low signal-to-noise ratio (SNR) conditions caused by thick liquid layers. Image quality deteriorates with increasing liquid thickness due to inelastic electron scattering. SAM~2's performance is further challenged by the stochastic dynamics of the nanoparticles, including large frame-to-frame displacements and transient departures from the field of view. Existing deep learning approaches for static TEM segmentation~\cite{yao_machine_2020,horwath2020understanding,sytwu2022understanding,kalinin2023machine,wang2021autodetect,cheng2023deep,rangel2024robust,ye2024diversitybased} likewise generalize poorly to LPTEM data and discard temporal context by processing frames independently.

To address these challenges, we introduce SAM-EM, a domain-adapted framework that extends SAM~2 through full model fine-tuning on 46,600 synthetic LPTEM video frames (Fig.~\ref{fig:fig1}). SAM-EM unifies segmentation, tracking, and statistical analysis into a single pipeline, integrating tools for mean-squared displacement (MSD), displacement distributions, and other relevant canonical statistical metric for spatiotemporal trajectories. Full fine-tuning enables SAM-EM to maintain segmentation fidelity under low-SNR conditions caused by thicker liquid samples, the most challenging yet also the most experimentally relevant regime for LPTEM. By combining robust video segmentation with quantitative analysis, SAM-EM moves LPTEM toward its envisioned role as a feedback-driven experimental platform for data-driven materials discovery.









\section{Methods}

\subsection{Dataset Generation}

To train and evaluate SAM-EM, we generated synthetic LPTEM videos with ground-truth masks following the method of Yao et al.\cite{yao_machine_2020}. The original code was modified to reflect experimental conditions (e.g., liquid thickness, particle size/shape, and electron beam dose rate). The particle positions were sampled from LEONARDO, a physics-informed variational autoencoder for stochastic diffusion in LPTEM~\cite{shabeeb2025learning}. The data was generated at 0.25 nm per pixel, corresponding to a 256 nm$\times$256 nm field of view (1024$\times$1024 pixels). By varying the simulated liquid thickness from 5 nm to 160 nm, we produced a dataset spanning a wide range of SNR values representative of experimental conditions (Appendix~\ref{appendix:a}).
The final dataset consisted of 1,000 synthetic videos, each 50 frames in length, with ground-truth masks and centroid positions. We reserved 68 videos for validation and testing, and additionally generated longer test sequences from independently sampled LEONARDO trajectories to assess generalization beyond 50 frames. Example frames are shown in Appendix~\ref{appendix:a}.

\subsection{SAM~2 Video Segmentation and Fine-tuning}

We fine-tuned SAM~2 for LPTEM video segmentation following the protocol of Ravi et al.~\cite{ravi2024sam}, using the synthetic datasets described in Section~\ref{appendix:a}. Of the 1,000 generated videos, 932 were used for training and the remainder reserved for validation and testing. Training examples included particle collisions and overlaps to expose the model to motion patterns characteristic of LPTEM data. Fine-tuning was performed using the official {\texttt{train.py}} script from the SAM~2 codebase, with the default model architecture and the training algorithm left unchanged. The primary configuration modification was to enforce box-prompt conditioning during training. Ground-truth masks were supplied as bounding box prompts, reflecting how users interact with the SAM-EM interface by drawing approximate boxes around particles. Consistent with the prior work in domain-adapted foundation models such as $\mu$SAM~\cite{archit2025}, we opted for full model fine-tuning rather than adapter-based methods, as it produced statistically superior $\mathcal{J}$\&$\mathcal{F}$ scores and qualitatively more stable masks and trajectories in our experiments. Additional details regarding our fine-tuning setup are provided in Appendix~\ref{appendix:b}.

\subsection{U-Net LPTEM Image Segmentation Baseline}
For benchmarking, we implemented the U-Net model from Yao et al.~\cite{yao_machine_2020}, which to our knowledge is the only deep learning–based segmentation model specifically developed for LPTEM data. We trained the model using the dataset provided by the authors for 20 epochs with a learning rate of $1\times10^{-6}$. Unlike SAM~2 and our fine-tuned SAM-EM framework, this U-Net model does not support prompt-based refinement, which limits its ability to adapt to complex or noisy scenarios. As a result, its segmentation performance was lower in our experiments. Nevertheless, U-Net has the advantage of efficiency; it produces predictions substantially faster and with lower computational cost compared to SAM~2 and SAM-EM.

\section{Experiments}

\subsection{Experimental Setup}

To evaluate the performance and robustness of SAM-EM, we constructed a testing set consisting of five distinct trials designed to capture a range of experimental conditions (details in Appendix~\ref{appendix:a}). The evaluation was performed on three standard segmentation and tracking metrics: Jaccard index (IoU), F1-score, and the composite $\mathcal{J}$\&$\mathcal{F}$ score. Formal definitions of these metrics are provided in Appendix~\ref{appendix:a1}.

\subsection{Results}

\setlength{\textfloatsep}{3em}
\begin{figure}[h!]
\centering
\includegraphics[width=0.6\textwidth]{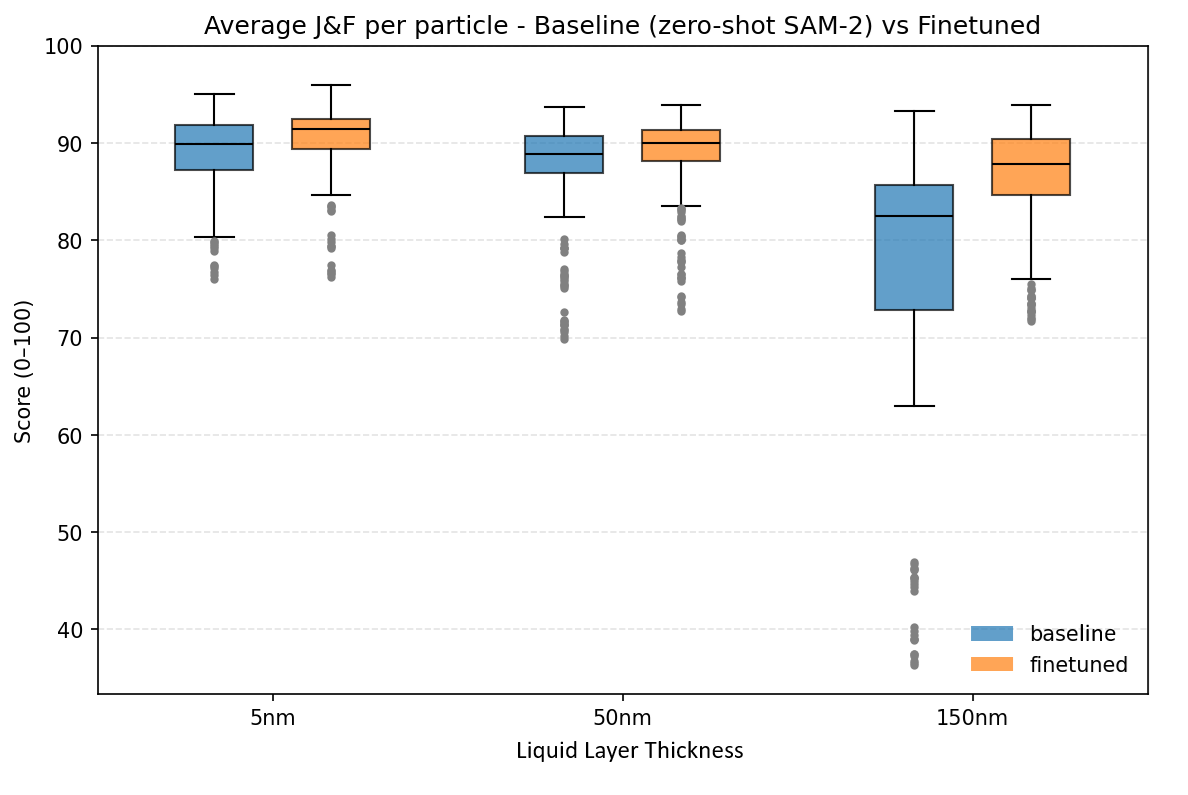}
\caption{{\bf Performance accuracy of SAM-EM in segmentation on simulated datasets.} Comparison of fine-tuned SAM-EM model and baseline (zero-shot SAM~2) on a simulated test dataset with 300 frames with two moving nanoparticles and with minimal particle interaction.}
\label{fig:fig2}
\end{figure}

Fig.~\ref{fig:fig2} shows SAM-EM consistently outperforms the zero-shot SAM~2 baseline on LPTEM segmentation as measured by the $\mathcal{J}$\&$\mathcal{F}$ score (higher, better). Improvements are modest for thin liquid layers (5 nm and 50 nm) but become pronounced under thicker liquid conditions, where fine-tuning yields a significant performance gain. This regime is particularly important because most commercial LPTEM chambers employ lithographically defined spacers of either 50 nm or 150 nm; the latter is frequently used as it imposes less confinement on the specimen. However, increased liquid thickness leads to additional electron scattering, lowering the image contrast and effective SNR, which makes segmentation more challenging. Strong performance in this regime therefore demonstrates both practical relevance and technical robustness.

The U-Net baseline achieved median $\mathcal{J}$\&$\mathcal{F}$ scores of approximately 0.5 across thicknesses, confirming that SAM~2 and SAM-EM substantially outperform existing LPTEM-specific segmentation approaches under the evaluated conditions.

Following segmentation, trajectory analysis on the extracted masks demonstrated reliable multi-particle tracking (Appendix~\ref{appendix:d}). The reconstructed particle paths are physically realistic, with minimal identity switching observed except under highly complex video conditions, where the video contains an unusually large number of particles that overlap in many of the frames. These results support the applicability of SAM-EM for tracking nanoparticle dynamics during self-assembly processes in LPTEM experiments.

\section{Conclusion}

We presented SAM-EM, a domain-adapted foundation model that unifies segmentation, tracking, and statistical analysis for LPTEM videos. Through full model fine-tuning of SAM~2 on curated synthetic datasets, SAM-EM achieved significant improvements in mask fidelity and temporal identity stability, with particularly strong gains under low SNR conditions caused by increased liquid sample thickness. Beyond segmentation accuracy, SAM-EM enables interactive selection of particles and immediate visualization of trajectory statistics such as mean-squared displacement and displacement distributions. This integration of robust segmentation with interpretable analysis opens a path toward closed-loop LPTEM workflows, where on-the-fly tracking can support automated experimentation. Such capabilities could ultimately allow researchers to probe defect dynamics in nanocrystals or follow the self-assembly of nanoparticles into ordered superlattices, directly connecting imaging to materials characterization and design. While our main evaluations use physics-grounded synthetic datasets, we also include qualitative results on real experimental videos and centroid-agreement analysis under noisy human-based annotation (Appendix~\ref{appendix:e}).

Although SAM-EM demonstrates strong performance, there are natural areas for further development. For example, distinguishing particles during severe overlap remains challenging, and model performance may vary across experimental setups due to differences in microscope conditions. However, these limitations are not fundamental. They can be addressed by expanding training datasets to include multi-lab experimental data. We view SAM-EM as an important first step toward a more generalizable framework and anticipate that community-driven data sharing and iterative refinement will further strengthen its utility across diverse in situ electron microscopy studies.

\section{Acknowledgment}
This research was supported by the NSF, Division of Chemical, Bioengineering, Environmental, and Transport Systems under award 2338466, Georgia Tech Institute for Matter and Systems, Exponential Electronics seed grant, the American Chemical Society Petroleum Research Fund under award 67239-DNI5, and the Exponential Electronics Seed grant of the Institute for Matter and Systems at Georgia Tech. The authors acknowledge the support of the Material Characterization Facility and the Electron Microscopy Facility of the Institute for Matter and Systems at Georgia Tech.

\bibliographystyle{plainnat}
\bibliography{main}

\newpage
\appendix

{\bf \large Appendix}

\section{Dataset and Simulation Details}
\label{appendix:a}

Our simulated training set consists of 932 videos each with 50 frames of size of 1024$\times$1024 pixels, where each of the videos has water thickness uniformly randomly sampled from a distribution of \(\{5,10,25,50,75,100,125,150,160\}\) nanometers (nm), with visual differences shown in Fig.~\ref{fig:fig5}. Each video contains a number of particles selected uniformly from a range of 1 to 8, with possible various interactions between particles, such as collisions or overlap, ensuring robustness of training.

For our testing set, we used 
\begin{enumerate}
    \item Video (1024$\times$1024) consisting of 300 frames and 3 particles, including occasional frame-to-frame overlaps, used for general performance evaluation (Fig.~\ref{fig:fig2})  
    \item Video (512$\times$512) consisting of 275 frames and a single particle,   providing a lower-resolution test case (Fig.~\ref{fig:fig4}) 
    \item Video (1024$\times$1024) consisting of 600 frames and 5 particles, with minimal interactions, used to evaluate tracking accuracy and mask stability over long trajectories (Fig.~\ref{fig:fig7}).
\end{enumerate}
\begin{figure*}[htbp!]
\centering
\includegraphics[width=\textwidth]{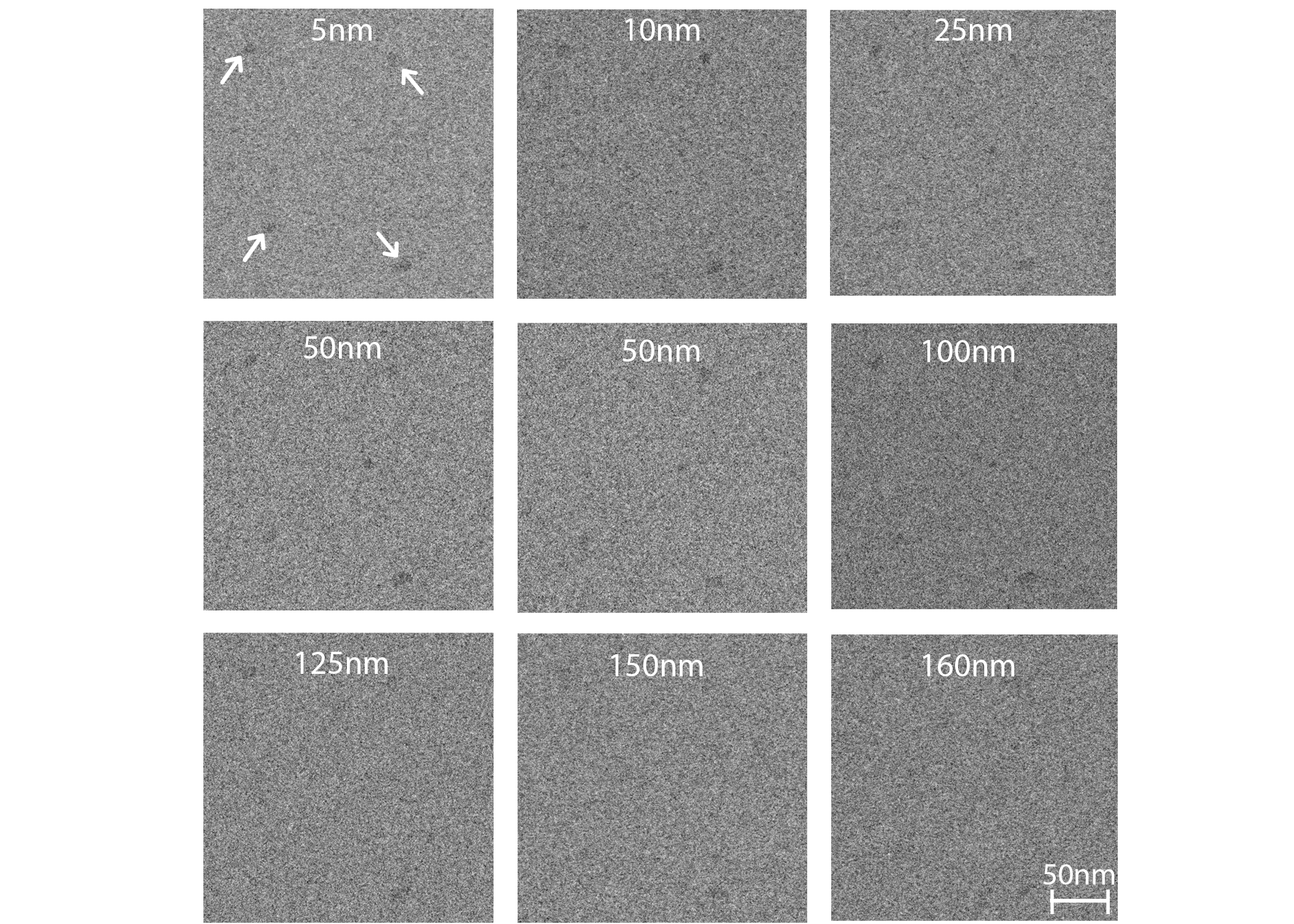}
\caption{Example simulated LPTEM frames at different liquid thicknesses, illustrating the decrease in image SNR as thickness (and electron scattering) increase.}
\label{fig:fig3}
\end{figure*}

\begin{figure*}[htbp!]
\centering
\includegraphics[width=\textwidth]{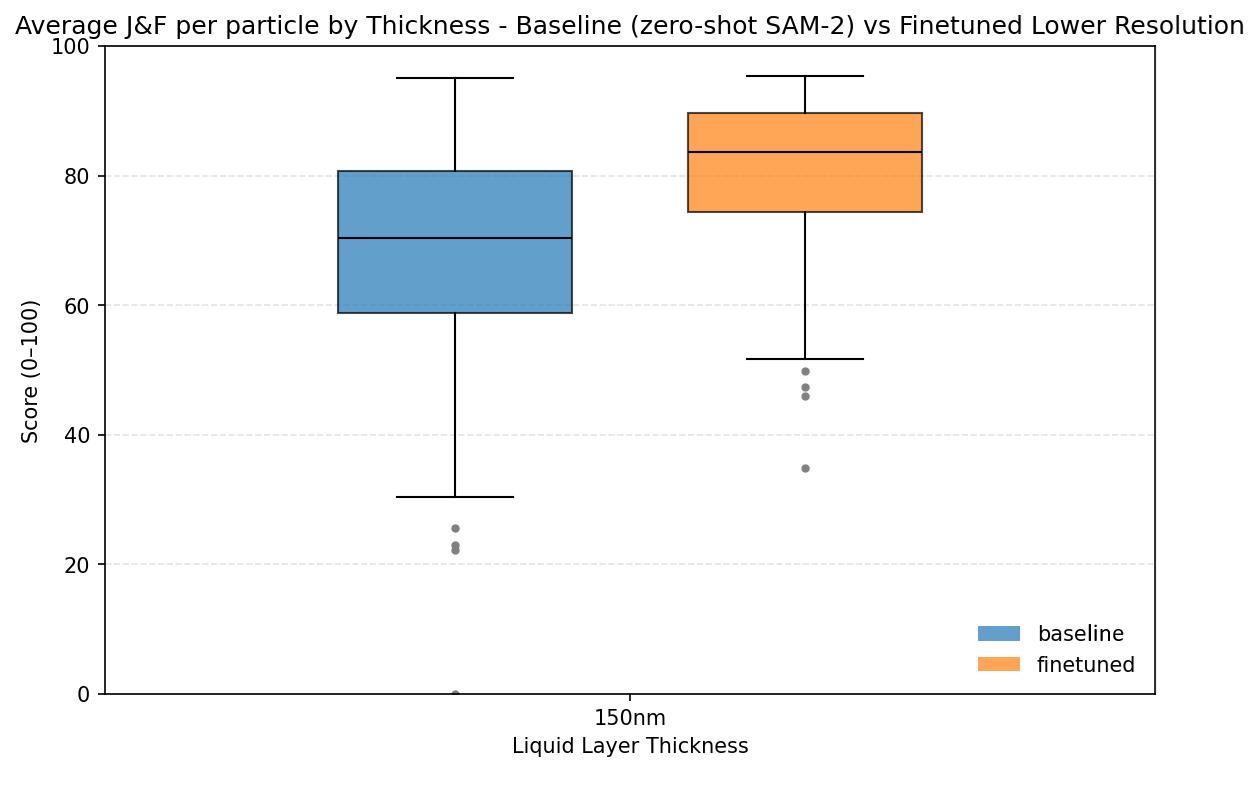}
\caption{Performance accuracy of SAM-EM in segmentation on a simulated dataset with resolution of 512$\times$512 resolution. Comparison of fine-tuned SAM-EM model and baseline (zero-shot SAM~2) on a simulated test dataset with 275 frames with one moving nanoparticle.}
\label{fig:fig4}
\end{figure*}

\begin{figure*}[htbp!]
\centering
\includegraphics[width=\textwidth]{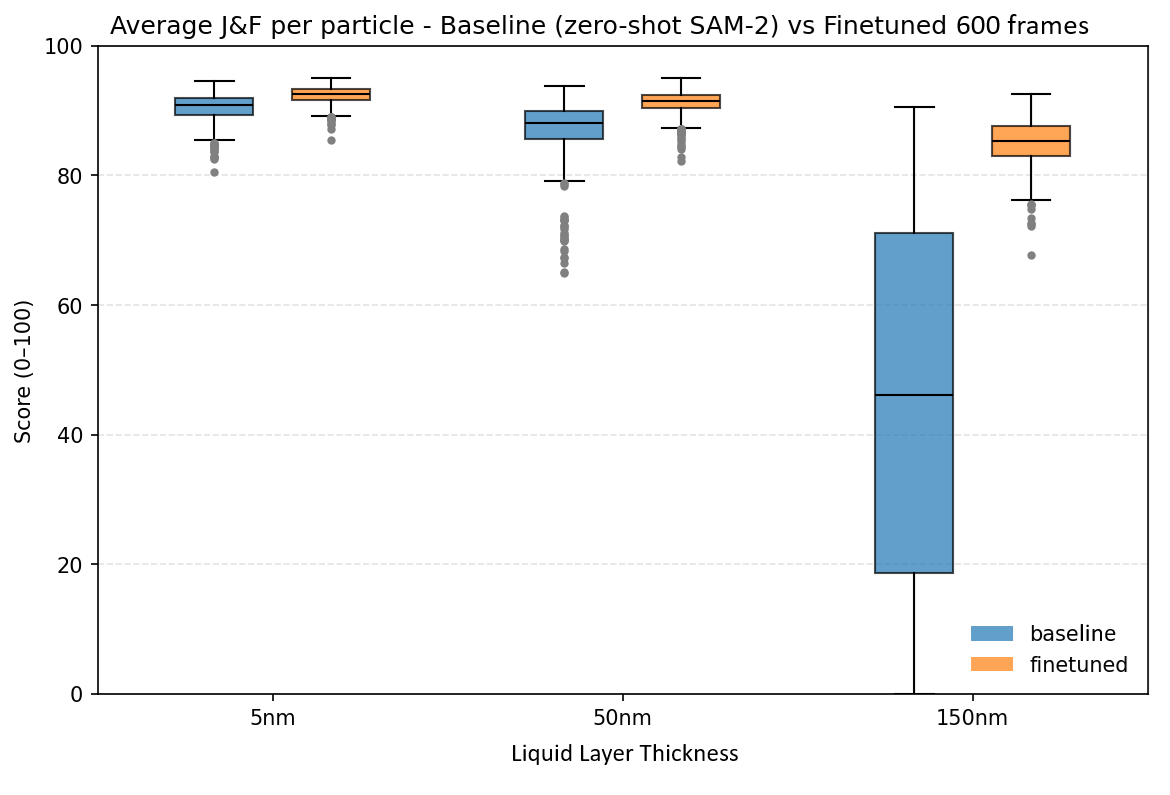}
\caption{Performance accuracy of SAM-EM in segmentation on simulated datasets with resolution of 1024$\times$1024. Comparison of fine-tuned SAM-EM model and baseline (zero-shot SAM~2) on a simulated test dataset with 600 frames with four moving nanoparticles and with minimal particle interaction.}
\label{fig:fig5}
\end{figure*}

\subsection{{Accuracy Metrics}}
\label{appendix:a1}
The accuracy of the segmentation on the simulated single-particle datasets was evaluated using the $\mathcal{J}$ \& $\mathcal{F}$ metric~\cite{ravi2024sam} on a scale from 0\% to 100\%, where a score of 100\% indicates that the prediction completely matches the ground truth. The Jaccard Index $\mathcal{J}$ assesses region similarity by calculating the intersection-over-union between the segmented mask ($M$) and the ground-truth ($G$): $\mathcal{J} = \frac{|M \cap G|}{|M \cup G|}$. The F-measure $\mathcal{F}$ assesses the accuracy of the mask contour. $\mathcal{F}$ is calculated from the contour-based precision $(P_c)$ and recall $(R_c)$ between the boundary pixels of the segmented mask and the ground-truth: $\mathcal{F} = \frac{2*P_c*R_c}{P_c+R_c}$. See \cite{perazzi2017} for further details on the $\mathcal{J}$ and $\mathcal{F}$ metrics. The $\mathcal{J}$ \& $\mathcal{F}$ score of a frame is the average of the $\mathcal{J}$ and $\mathcal{F}$ of the frame~\cite{pont20172017}. The average $\mathcal{J}$ \& $\mathcal{F}$ across the video is computed using the per-frame scores.

\section{Fine-tuning Details}
\label{appendix:b}

We fine-tuned the SAM~2 Hiera-Large checkpoint on synthetic LPTEM video datasets using a single NVIDIA GH200 GPU (96 GB VRAM). Training used a batch size of 1 (constrained by memory), a vision learning rate of \(1\times10^{-7}\), a frame count of 8 (i.e., the number of consecutive frames the model was exposed to per step), and 15 epochs. These hyperparameters were selected empirically to balance stability and convergence for noisy video data. The full YAML configuration, along with training scripts, is available in our public GitHub repository.

\section{Fine-tuning Loss Curves}

\begin{figure*}[htbp!]
\centering
\includegraphics[scale = 0.7]{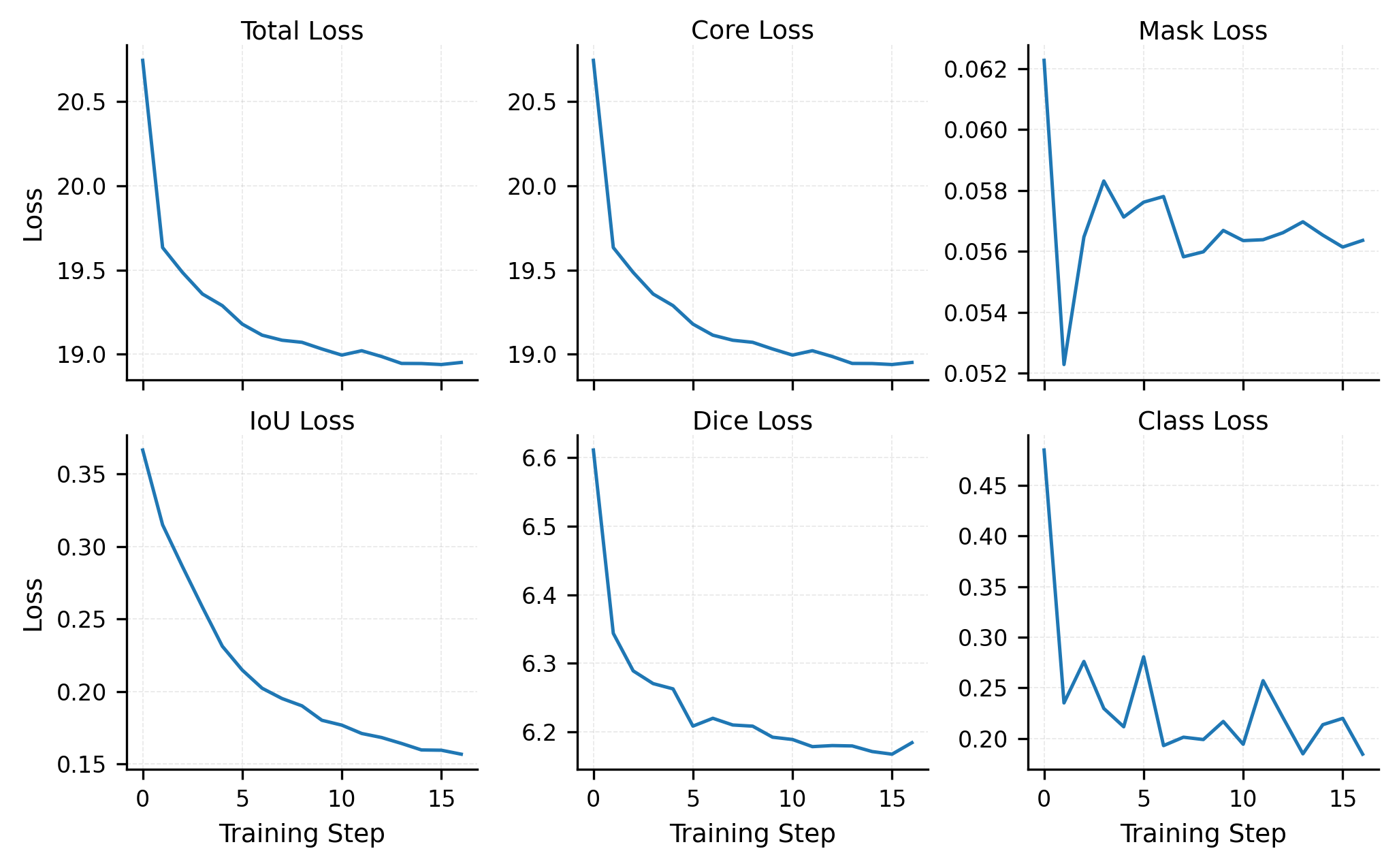}
\caption{{\bf Generated training loss curves during model fine-tuning}. The higher initial decline and gradual slowing of the total loss show a healthy loss curve, suggesting an appropriate learning rate}
\label{fig:fig6}
\end{figure*}

\section{Statistical Analysis of Particle Trajectories}
\label{appendix:d}
\subsection{Single particle trajectory analysis}
The centroid (\(x\), \(y\)) and orientation (in-plane angle, $\theta$) of each particle in each frame are calculated from the segmented mask array in the Particle Tracking module. The data is saved as a .csv file.

To characterize the diffusive behavior of the nanoparticles in LPTEM experiments, we analyzed the trajectories using three key statistical metrics.
\subsubsection{Time-averaged Mean Squared Displacement}
The time-averaged $\text{MSD}(\tau)$ quantifies the average displacement of a particle over time delay, $\tau$. It is given by:

\begin{equation*}
\text{MSD}(\tau) = \langle (\mathbf{r}(t + \tau) - \mathbf{r}(t))^2 \rangle,
\end{equation*}

where $\mathbf{r}(t)$ and $\mathbf{r}(t + \tau)$ represent the particle's position at time $t$ and $t+\tau$, respectively. $\langle \cdot \rangle$ denotes the mean value over frames of a single trajectory.

\subsubsection{Velocity Autocorrelation}
The velocity autocorrelation function, $C_\mathbf{v}(\tau)$, measures the temporal memory of a particle’s motion and is defined as:
\begin{equation*}
C_\mathbf{v}(\tau) = \frac{\langle \mathbf{v}(t) \cdot \mathbf{v}(t + \tau) \rangle}{\langle \mathbf{v}(t)^2 \rangle},
\end{equation*}
where $\mathbf{v}(t)$ and $\mathbf{v}(t + \tau)$ represent the particle's velocities at time $t$ and $t+\tau$, respectively. This function indicates whether the motion is correlated, uncorrelated, or anti-correlated.

\subsubsection{Distribution of Displacements}
The distribution of displacements provides statistical information about the nature of the particle’s motion. This is characterized by the probability density function (PDF) of particle displacements $\Delta \mathbf{r}$ over a fixed time interval. For Brownian motion, the distribution is Gaussian, while non-Gaussian processes may exhibit heavy tails or other deviations, reflecting the underlying physical mechanisms influencing the motion.
\begin{figure}[h!]
\centering
\includegraphics[width = \textwidth]{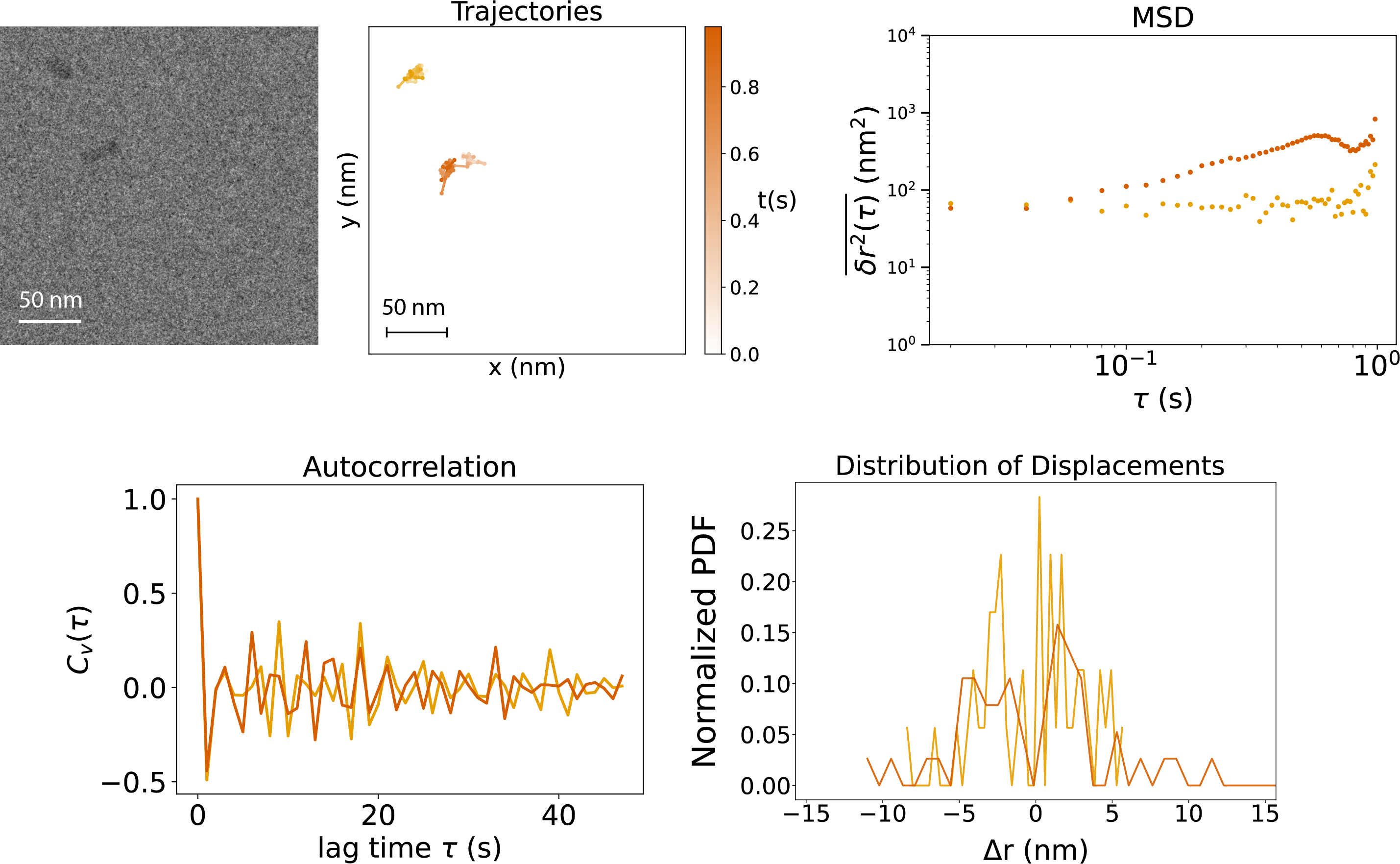}
\caption{Statistical analysis and trajectory plots of two particles in high-noise LPTEM simulated data (50 frames, 5nm liquid layer), demonstrating that SAM-EM maintains accurate tracking and quantitative characterization (MSD, distribution of displacement, and velocity autocorrelation) under low-SNR conditions.}
\label{fig:fig7}
\end{figure}

\section{Real Experimental Data Evaluation}
\label{appendix:e}
\subsection{Qualitative Analysis}
\begin{figure}[h!]
\centering
\includegraphics[width = \textwidth]{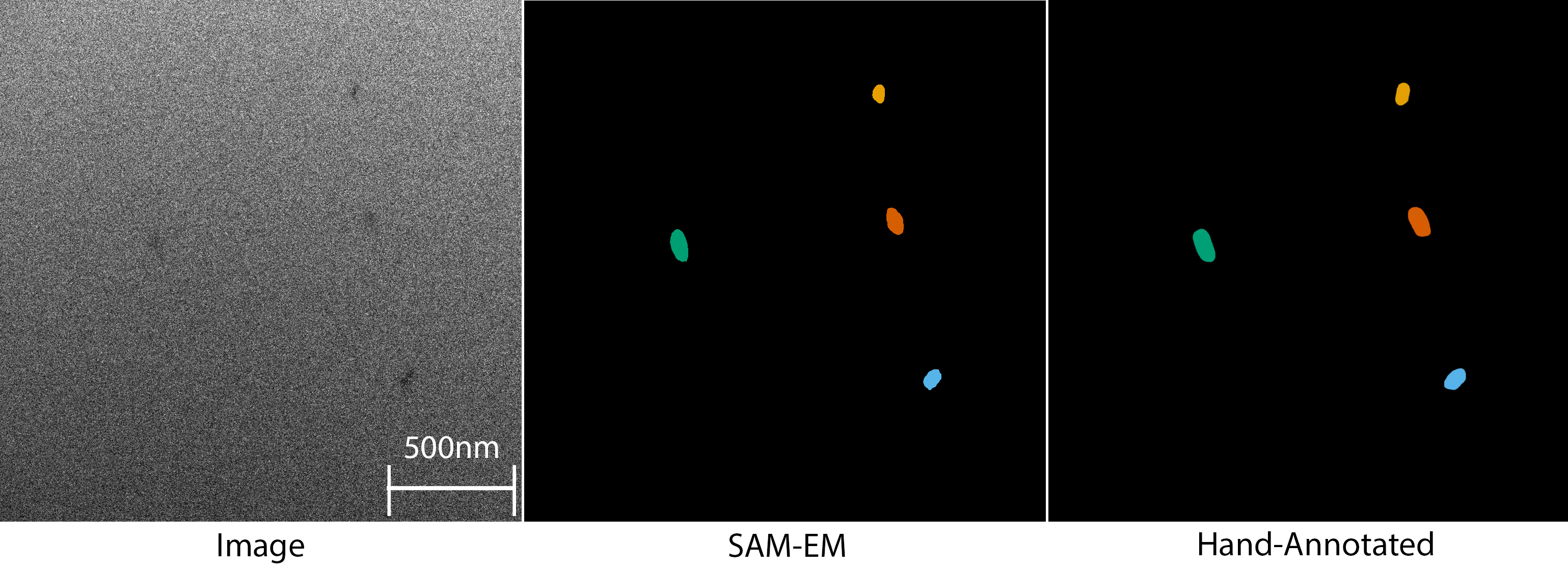}
\caption{Real LPTEM frame with hand-annotated labels and SAM-EM segmentation masks.
Left: raw image. Middle: SAM-EM segmentation (one box prompt per particle). Right: hand-annotated masks used as \emph{pseudo}-ground truth for visualization. SAM-EM identifies the same four nanoparticles with compact, identity-consistent masks. Apparent boundary differences largely reflect label imprecision under low SNR rather than missed detections.}
\label{fig:fig8}
\end{figure}
To showcase the power of SAM-EM on real LPTEM data, we applied it on an experimental LPTEM video (200 frames) acquired at an electron dose rate of 35 e$^-$/\AA$^2\cdot$s and at a frame rate of 80 fps, equivalent to 0.0125 seconds exposure time. For this low-contrast dataset, SAM-EM generates coherent ellipsoidal masks that align closely with manually annotated particle masks. The largest discrepancies appear along particle boundaries, where hand-annotated masks tend to be slightly thicker or rotated, an expected deviation given both the human annotation error and the weak signal at nanoparticle edges. Visual inspection confirms that centroid positions remain well-aligned across all four particles, even when mask sizes differ.

\subsection{Centroid Agreement Analysis}
\begin{figure}[h!]
\centering
\includegraphics[width = 0.5\textwidth]{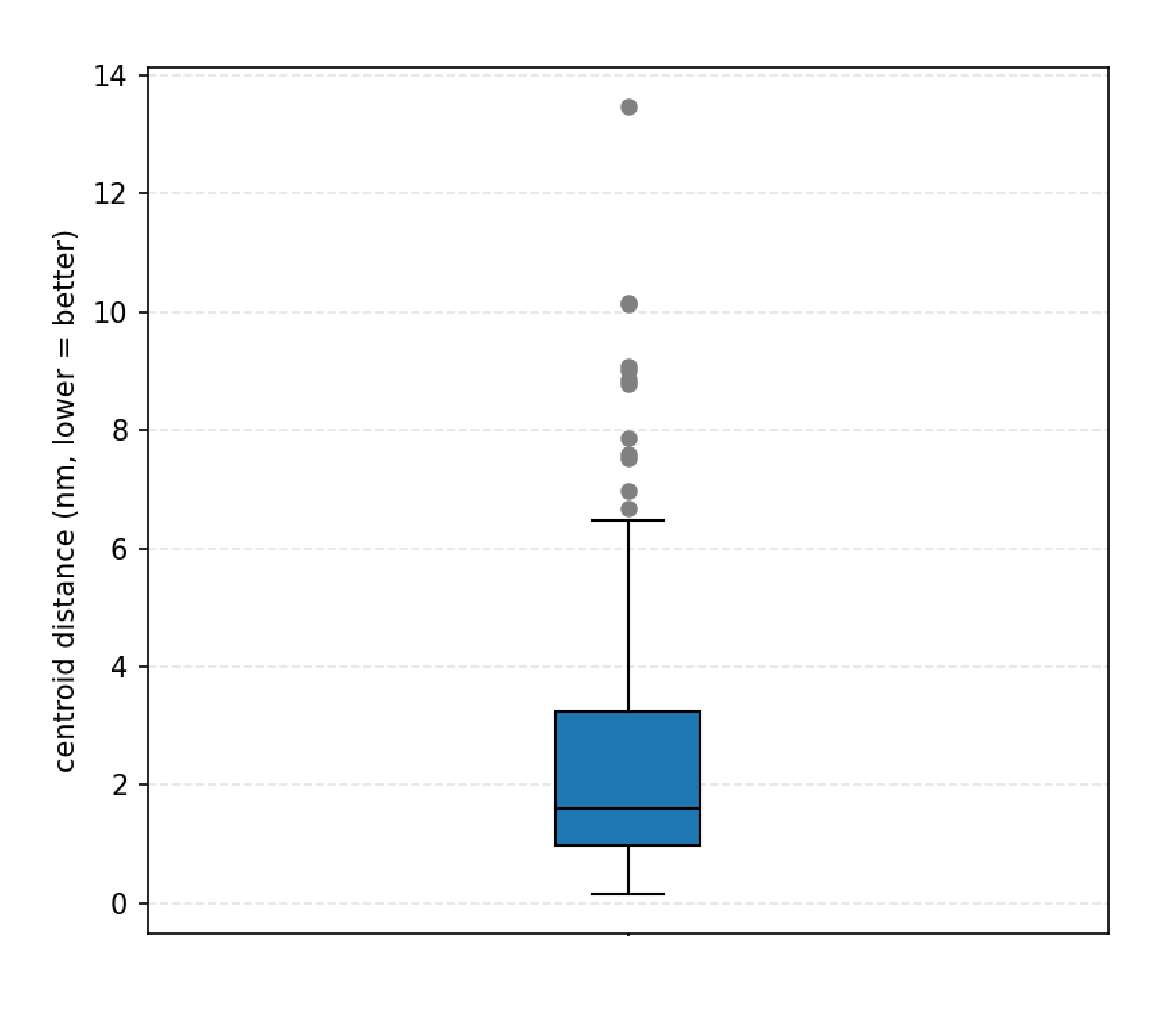}
\caption{Box plot of centroid distances (nm) between SAM-EM predictions and manually annotated particle centers across 200 frames from a real LPTEM dataset. The median offset of ~1.8 nm is within the expected uncertainty of manual annotation under low-SNR imaging conditions.}
\label{fig:fig9}
\end{figure}
Because hand-annotated data in LPTEM is inherently noisy and inconsistent at the pixel level, we evaluated SAM-EM using centroid agreement rather than mask accuracy. Under low SNR conditions, the nanoparticle boundaries are often subjective and can be interpreted differently. Different annotators may draw slightly thicker or rotated masks even when identifying the same particle, leading to misleading results if mask accuracy is used directly for evaluation. As shown in Fig.~\ref{fig:fig9}, SAM-EM achieves a median centroid deviation of $\approx$1.8 nm relative to hand-annotated masks. The narrow interquartile range indicates consistent alignment between model predictions and manual annotations, while the presence of outliers reflects regions where boundaries are highly occluded or ambiguous. This analysis demonstrates that SAM-EM reliably localizes particle centers with sub-particle precision, making it suitable for downstream single-particle trajectory analysis and quantitative studies of nanoscale motion. We note that, unlike the simulated data in previous figures, experimental datasets lack a true ground truth; thus, Fig.~\ref{fig:fig9} reports only the distance between centroids in SAM-EM and manually segmented masks.
\section{Resources}
\label{appendix:f}
All of our code is available on GitHub: 
\href{https://github.com/JamaliLab/SAM-EM}{github.com/JamaliLab/SAM-EM}.

Code for simulated data generation:
\href{https://github.com/JamaliLab/LPTEMsimulator}{github.com/JamaliLab/LPTEMsimulator}.

All datasets and model checkpoints are available on Hugging Face: \href{https://huggingface.co/sam-em-paper}{https://huggingface.co/sam-em-paper}

\newpage

\end{document}